# High-speed Railway Fastener Detection and Localization Method based on convolutional neural network

Qing Song[1], Yao Guo[1], Jianan Jiang[1], Chun Liu[1], Mengjie Hu[1]

*1.Pattern Recognition and Intelligent Vision Lab, Beijing University of Posts and Telecommunications, Beijing, China*
*E-mail：priv@bupt.edu.cn*

**Abstract:**
Railway transportation is the artery of China's national economy and plays an important role in the development of today's society. Due to the late start of China's railway security inspection technology, the current railway security inspection tasks mainly rely on manual inspection, but the manual inspection efficiency is low, and a lot of manpower and material resources are needed. In this paper, we establish a steel rail fastener detection image dataset, which contains 4,000 rail fastener pictures about 4 types. We use the regional suggestion network to generate the region of interest, extracts the features using the convolutional neural network, and fuses the classifier into the detection network. With online hard sample mining to improve the accuracy of the model, we optimize the Faster RCNN detection framework by reducing the number of regions of interest. Finally, the model accuracy reaches 99% and the speed reaches 35FPS in the deployment environment of TITAN X GPU.

## 1. Introduction

Rail fastener is a part used to connect rail sleeper and the rail, maintain gauge and prevent the longitudinal and transverse deviation of rail from sleeper. If the rail fastener is missing, it will cause the rail distance offset. It will cause worsening wheel flange wear, structural deformation, construction cracks. What's more, it will cause train derailment. Therefore, the daily inspection and maintenance of the rail fastener is of great significance in the safety of railway transportation.

At present, there are two main ways of railway patrol inspection. One is to train workers to conduct regular track patrol inspection to check whether there are rail fasteners falling off. This method requires a lot of manpower costs. Long-term outdoor inspection will affect the health of workers. Some routes are not suitable for manual inspection because of the bad environment. Another way is to use a patrol car equipped with a camera to take pictures of the rails at night, collect the video images of the railways, and then the trained workers will check whether the video images have rail fasteners falling off. So, fastener detection technology based on convolutional neural network and deep learning is drawing more and more attention.

Singh et al. [1] proposed a high-performance automatic track detection algorithm which combines image processing and analysis methods, with applying edge density localization to the detection of missing clips. The paper addressed the issues of finding missing clips and finding blue clips which have been recently replaced in place of damaged clips.This method detects blue clips based on the color information in the located windows. The former references pioneered a series of methods to locate fasteners according to the existing geometric position relations between railway sections. References [2][3] detect missing fasteners using a patented sensor devoted to the high precision 3D reconstruction, which is based on a MLP neural network classifier using wavelets. References [4] first adopted BP neural network and principal component analysis for automatically detecting the absence of the fastening bolts that secure the rails to the sleepers.Xia et al. [5] came up with a from-coarse-to-fine algorithm according to the Haar-like feature set according to the fastener's geometrical characteristics. By dividing into four parts and training independently with a highly effective multiple classifier, Adaboost, they gained a satisfactory detection rate. Liu et al. [6] used sparse representation algorithm and directional gradient pyramid histogram (PHOG) to identify two symmetrical fasteners. Similarly, by extracting orientation fields as feature descriptions of fasteners, Yang et al. [7] used Direction Field extracted as a stable feature descriptor and orientation coherence for matching the fasteners. Li et al. [8] extracted edge features to locate the fastener region. Feng et al. [9] apply probabilistic topic model to extract Haar-like features and locate the fastener region in order to detect partially destroyed and completely missing fasteners.

However, these algorithms are based on traditional manual feature extraction, such as hog, lbp, edge and haar, and most of them are not very good in detection effect, lacking generalization ability for illumination changes and weather changes. In addition, because the extracted Features do not adequately characterize their critical components, the positioning accuracy of the fasteners is low.

In this paper, We propose an algorithm for high-speed railway fastener detection and localization based on convolutional neural network, which can effectively detect the railway fasteners under different situations. We attempt to add the online hard example mining method to our algorithm based on Faster R-CNN[13] detection framework. Also, in this paper, we adjust the details of Faster R-CNN network in order to accelerate the method. The proposed method is robust to almost every complicated situations faced in the high-speed railway system. Due to region limitations, this method considers four different common types of fastener used in high speed railways.

With the above changes proposed, our contribution can be summarized as below:

1. We proposed a simple yet effective railway fastener detection based on CNNs which can offer a safe but accurate method compared with traditional detection method.

2. We construct a dataset for fastener detection with containing four different types of railway fastener in different conditions.

3. According to the task, we attempt the online hard example mining model and adjust the region proposal method for a better performance on both accuracy and speed.

## 2. Related Work

The research of image object recognition has a history of more than 50 years, and various theories and algorithms emerge in endlessly. The object detection task appeared relatively late because of its complexity. We summarized the main progress of object classification and detection algorithms in recent years based on the international visual competition PASCAL VOC. This series of competitions had a profound impact on the development of object classification and detection, and its work also represented the highest level at that time.

PASCAL VOC competition has introduced object detection task contest since the first session of 2005. The main task is to predict the object categories and external rectangular boxes contained in the test pictures. The most important difference between object detection task and object classification task is that object structure information plays an important role in object detection, while object classification is more concerned with the global representation of objects or images. The input of object detection is a window containing objects, while object classification is the whole image. For a given window, object classification and object detection are very similar in feature extraction, feature coding and classifier design. In the first stage of the development of image detection, all methods use traditional methods. Although there are many innovations and breakthroughs, the overall effect is general, and no ideal results have been achieved. With the development of deep learning, deep learning is gradually applied to image detection, which is the second stage of image detection. The new method put forward, leaving traditional methods far behind. With the emergence of new methods, Mu Biao Jian has made considerable progress. Here are some representative methods.

In 2014, RCNN [11], which was proposed by Ross Girshick, a Google scientist, obtains about 2,000 candidate regions in the image by selective search method, extracts deep convolution features from each candidate region, and trains SVM classifier with these features to obtain the deep convolution features. For each candidate region, non-maximum suppression is used to obtain the precise boundaries of each object. The average accuracy of DPM HSC in PASCAL VOC2007 dataset has nearly doubled (from 34.3% to 66%). However, the steps of this method are tedious and the detection speed is very slow. At present, it has been replaced by a new method, Fast RCNN [12], which is an improvement on RCNN method. Fast RCNN proposed ROI pooling method to greatly reduce the computational redundancy of RCNN. All candidate regions only need one convolution operation of convolution neural network. SoftMax is used to replace SVM in image feature classification. Meanwhile, multi-task loss function is used to carry out border regression, which reduces the steps of RCNN and improves the detection accuracy. Later Google scientist Ross Girshick and others produced Faster RCNN [13], which is an improvement of RCNN and Fast RCNN methods. This method will reduce the difficulty and time-consuming of obtaining candidate regions, further reduce the time and complexity of image detection, and slightly improve the detection accuracy. R-FCN [14] can be regarded as an accelerated version of Faster RCNN. Before and after ensuring the accuracy fluctuation is not too large, the latter layers of calculation of convolution neural network are shared, which further reduces the image detection time, and the detection accuracy is slightly improved by combining deeper convolution network and online hard sample mining. At present, R-FCN is a very frontier research achievement, which achieves good results in the speed and accuracy of detection, and has a very high application prospect. As for the one-stage-detection, SSD[15], which shows a great balance between accuracy and speed, is based on position regression for depth learning image detection. Therefore, the detection speed of this method is very fast, and video detection can be realized on Titan X GPU. This method uses multi-scale Anchor mechanism to regress the multi-scale features of each position in the image, which not only ensures the detection speed, but also achieves the detection accuracy close to that of candidate region method.

Object detection technology based on deep learning is the mainstream solution of object detection. In a few years, considerable achievements have been achieved. Existing algorithms have been constantly improved and new algorithms have been continuously put out. It has achieved fairly good results. It has been widely studied and applied in academia and industry. The five detection methods mentioned above are the frontiers of research. Fruit. In the future, the use of deep learning to process image detection tasks is still a research hotspot, there will be more and better algorithms. Therefore, the research of rail fastener detection algorithm based on Faster RCNN has a very broad research prospects, which is in line with the development trend of national and social science and technology.

## 3. Method

### 3.1 Our Dataset and pre-process

Image detection technology based on deep learning is a supervised machine learning method. At present, there is no free and open source image database of rail fasteners for this project, so image acquisition of rail fasteners is needed. In this scheme, rail fasteners are classified into four categories: W300-1 fasteners, WJ-7 fasteners, WJ-8 fasteners and V fasteners. The examples of each fastener data is shown in Fig.1. For each type of fasteners, image data of multi-scale, multi-angle, multi-illumination conditions and multi-model need to be collected. The more sufficient the data, the better the data, and updated with the types of rail fasteners.

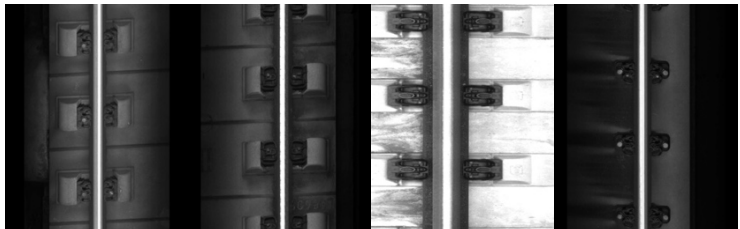

Figure 1. Four different categories of railway fastener in the dataset.

The four types of fastener pictures collected are preprocessed as follow: because the size of the original picture is too large, the original picture needs to be scaled. It is noted that the distribution of rail fasteners is in accordance with the longitudinal distribution, so the image of rail fastener is scaled to 1000 according to the vertical height, the width is scaled to 800 according to the equal proportion, and the image with the width of more than 800 is scaled to the center line of the vertical axis. Cut the image of the rail fastener whose width is less than 800 after zooming, and fill the black edge on both sides. Tab.1 shows the separation of our dataset and the scales of the collected data.

Table 1. THE SEPARATION AND THE SCALES OF DATASET

| Dataset | Amount | Scale |
| --- | --- | --- |
| Train | 3200 | 800*1000 |
| Val | 800 | 800*1000 |

### 3.2 Backbone based on RPN and improvement

Sliding window method is similar to exhaustive image sub-region search, but in general, most of the sub-regions in the image have no objects. Scholars naturally think of improving computational efficiency by searching only the areas most likely to contain objects in an image. Selective search method is the most famous rectangular box extraction algorithm at present, which was proposed by Koen E.A in 2011. The following is a general description of the selection search algorithm: the main idea of the selection search algorithm is that the possible regions of the objects in the image should have some similarity or continuity. Therefore, the selection search is based on the above idea, and the sub-region merging method is used to extract candidate boundary boxes. Firstly, the input image segmentation algorithm generates many small sub-regions. Secondly, according to the similarity between these sub-regions (the similarity criteria are mainly color, texture, size, etc.), regions are merged and iterated. During each iteration, boundary boxes are created for these merged sub-regions as candidate boxes. We provide an intuitive representation which shows the diversity of sliding windows and region proposal method in Fig.2.

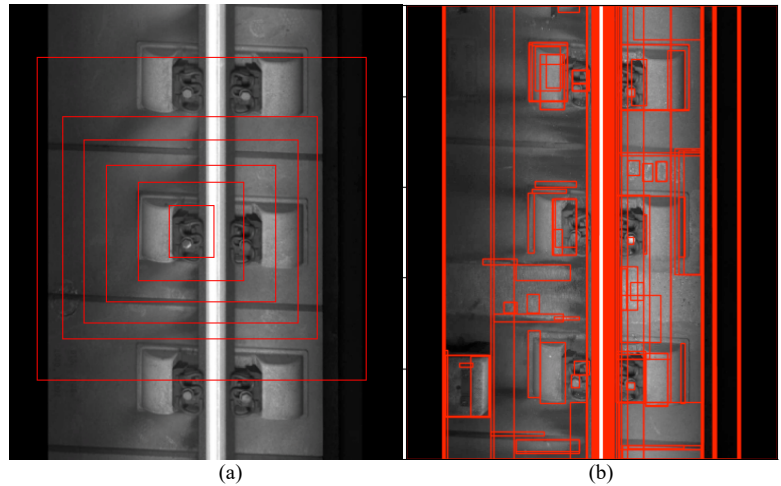

(a)                                            (b)

Figure 2. Comparison of (a)sliding windows and (b)region proposal method.

The regional recommendation method used in R-CNN and Fast R-CNN is selective search, which is separated from the subsequent classification algorithm and the previous feature extraction method. It consumes a lot of memory and has low computational efficiency. Faster R-CNN proposes Regional Proposal Network (RPN), which is connected with the feature extraction network and the classification network to achieve end-to-end detection network. One of the reasons why Faster R-CNN is faster than Fast R-CNN is that the regional recommendation network consists of full convolution layer.

RPN generates a feature vector of 256 or 512 dimensions by adding a 3x3 convolution layer behind the feature extraction network. Then the feature vectors are fed into two branches of the convolution layer. One convolution layer is used to predict the offset of anchor, and the other convolution layer is used to classify anchor. Anchors literally can be understood as an anchor, located at the centre of the sliding window of NxN mentioned earlier. For a sliding window, multiple proposals can be predicted simultaneously, assuming that there are k proposals. K proposals are k reference boxes, and each reference box can be uniquely determined by a scale, an aspect ratio and an anchor in sliding windows. Therefore, the anchor mentioned later in this article can be understood as an anchor box or a reference box. In this paper, the author defines k = 9, that is, three scales and three aspect ratios to determine the corresponding nine reference boxes at the current sliding window location, the output of 4xk regression layers and the output of 2xk classification layers. For a feature map of WxH, there are corresponding WxHxk anchors. All anchors have scale invariance.

Before calculating the loss function, it is necessary to set the positive and negative sample rules for anchors:

(1) If the IOU values of reference box and ground true corresponding to anchor are the largest, the positive samples are marked.

(2) If the IOU of reference box and ground truth corresponding to anchor is more than 0.7, it is marked as positive sample. In fact, using the second rule can basically find enough positive samples, but for some extreme cases, such as all anchor corresponding reference box and ground-truth IOU is not more than 0.7, the first rule can be used to generate.

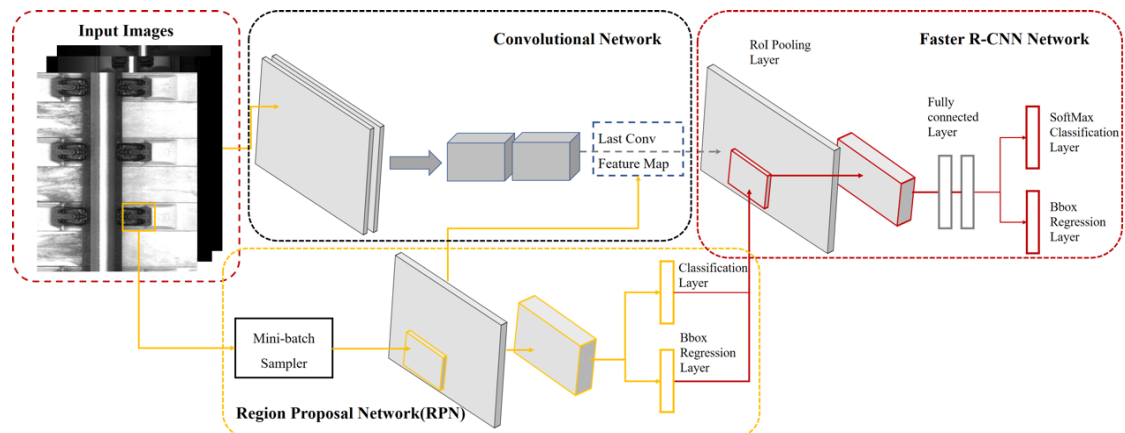

Figure 3. The Faster-RCNN network of fastener detection

RCNN module classifies and regresses the regions of interest output from Proposal layer, so the time-consuming part of RCNN is proportional to the number of regions of interest. Since the maximum number of rail fasteners in each picture is only 6, in this paper, we reduce RCNN time-consuming by reducing the number of regions of interest output from Proposal layer. The number of ROIs output by Proposal layer is 300. By reducing the number of ROIs to 50, the speed of single card model is increased to 31 FPS with the recall rate and precision rate reduced by 1%.

Furthermore, in order to reduce the computational complexity of RCNN module network and further improve the speed of model, the proposed area network can be moved from the sampling position in the fourth stage of Resnet 18 network to the sampling position in the fifth stage of Resnet network, so that the convolution layer of classification module can be reduced. At the same time, in order to ensure the size of feature graph, the convolution parameters in the fifth stage of Resnet 18 network can be reset without down-sampling. After ROI-Pooling, the full connection layer learning feature is directly used for classification and regression. Because of the role of online hard sample mining module, the accuracy of the model does not deteriorate. Through this operation, the model speed reaches 35 FPS.

### 3.3 Online Hard Example Mining Model

Hard samples refer to positive samples or negative samples of easily misclassified categories. V-type and WJ-7 rail fasteners are easy to be mis-detected and belong to hard samples. Online hard sample detection algorithm training is a region-based target detection algorithm, which can suppress easily distinguished samples and a small number of samples, so that the model can effectively train hard samples. The operation process of online hard example mining (OHEM) module can be illustrated in Fig.4.

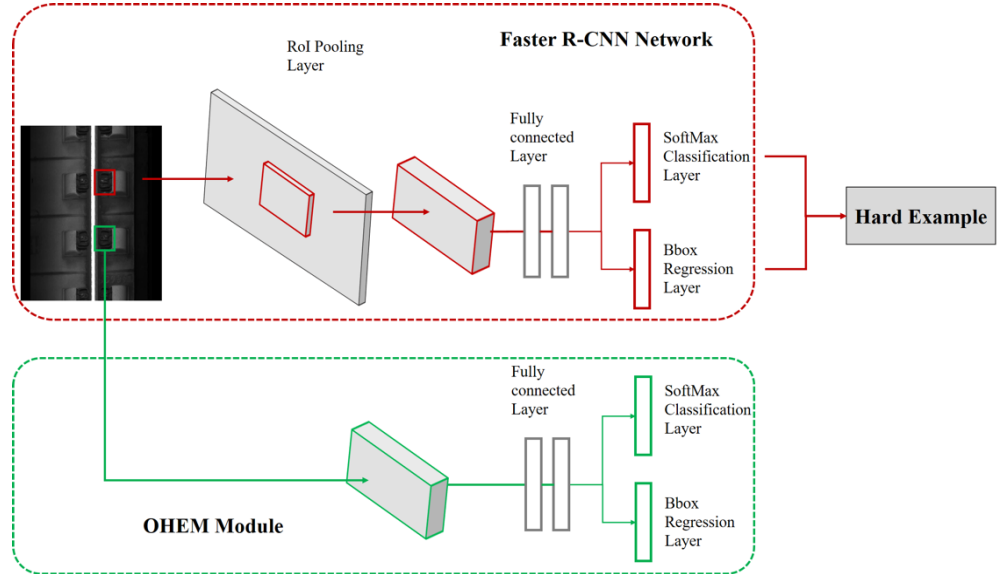

Figure 4. Online hard example mining module

The improved network has two identical ROI Pooling with RCNN networks, among which network.1 module is read-only. All the regions of interest output from the proposed network are computed forward once. In this stage, only the memory of forward transmission is allocated, the loss of each region of interest is calculated, and the loss values are sorted from large to small, then the first 256 interested regions are selected. Area feeding into network.2 module, in network.2 module, not only forward network calculation, but also backward propagation calculation, network weight updating, network.1 module and network.2 module share network weight. In the testing stage, only network.1 module is needed, which does not affect the testing speed of the network model.

## 4. Experiment

Image detection output target position and target category, intersection-over-union (IOU) is an index to measure the quality of target position. It represents the overlap rate between the predicted target location and the real labeled target location, i.e. the ratio of the intersection and union of the predicted target location and the real labeled target location. The formula is shown as follows (1):

$$\text{IOU} = \frac{PredictResult \cap GroundTruth}{PredictResult \cup GroundTruth} \qquad (1)$$

Through this index, we can judge whether the prediction result is good or bad. If the intersection ratio between the prediction box and the real labeled target exceeds the set threshold, we can judge that the prediction result is correct. If the prediction result is lower than the threshold, it means the model mis-detect the target. In this paper, the threshold of IOU is chosen as 0.75.

Meanwhile, IOU is used to measure whether a single model prediction box meets the requirement, and to judge the overall prediction results of the model, how many of the prediction results are correct and how many are missing need to be measured by Precision and Recall indicators. The accuracy rate is for the prediction results of the model. It shows how many of the prediction results are correct. There are two situations: one is to predict the original positive samples correctly, that is, the real class, and the other is to forecast the original negative samples incorrectly, that is, false positive examples. In image detection tasks, the accuracy rate represents the correct classification results, and the ratio of the intersection of the predicted position of the model output and the real labeled target position to the proportion of the target exceeding the threshold to the positive sample of the model predicted result. The formula of precision is shown as (2):

$$\text{Precision} = \frac{TruePositive}{TruePositive + FalsePositive} \qquad (2)$$

Also, the recall rate refers to the original sample. It shows how many positive samples are predicted correctly. There are two situations: one is that the original positive sample is predicted correctly, that is, the real class, and the other is that the original positive sample is predicted incorrectly, that is, the false counterexample. In image detection, the recall rate represents the correct classification result, and the intersection ratio between the predicted position of model output and the real target position is higher than the ratio of the target exceeding the threshold to the total number of targets.

And the recall is calculated as formula (3):

$$\text{Recall} = \frac{TruePositive}{TruePositive + FalseNegative} \qquad (3)$$

The aim of our study is to improve the accuracy as much as possible on the basis of guaranteeing the recall rate.

We use VGG16 Network as our backbone network, and the detection result of precision and recall with applying the VGG network is shown in Tab.2.

Table 2. Detection results with VGG16 backbone

| Category | Precision | Recall |
|---|---|---|
| V | 90.1% | 90.4% |
| W300-1 | 95.51% | 95.31% |
| WJ-7 | 91.40% | 89.55% |
| WJ-8 | 95.53% | 95.95% |
| mean | 93.14% | 92.80% |

Although the average accuracy and recall rate of the model are more than 90%, it still needs to be improved, especially for V and WJ-7 fasteners. There are a lot of missing fasteners. So, we change our backbone with ResNet18 which use skip connections to avoid gradient disappearing during training process. This residual network is easier to converge. The detection results with ResNet18 as backbone and with OHEM method are respectively shown in Tab.3 and Tab.4.

Table 3. Detection results with ResNet18 backbone

| Category | Precision | Recall |
|---|---|---|
| V | 94.1% | 94.4% |
| W300-1 | 99.51% | 99.31% |
| WJ-7 | 95.40% | 93.55% |
| WJ-8 | 99.53% | 99.95% |
| mean | 97.14% | 96.30% |

Table 4. Detection results with ResNet18 backbone and OHEM

| Category | Precision | Recall |
|---|---|---|
| V | 99.91% | 99.94% |
| W300-1 | 99.95% | 99.91% |
| WJ-7 | 99.53% | 99.95% |
| WJ-8 | 99.94% | 99.95% |
| mean | 99.83% | 99.93% |

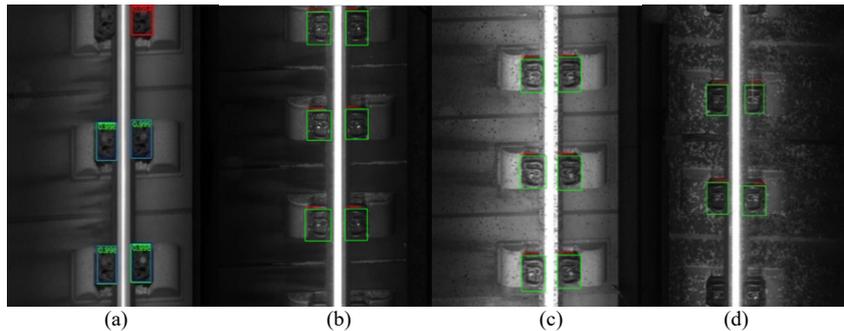

(a) (b) (c) (d)
Figure 5. (a)The missed detection and false detection of fasteners. (b)(c)(d) are the detection results with OHEM module

The detection network of rail fasteners after on-line difficult sample mining has greatly improved the detection accuracy of V and WJ-7 rail fasteners by fastener detection model without affecting the accuracy of W300-1 fasteners and WJ-8 fasteners. We visualize the detection results and Fig.5 shows the robust generalization ability of our method in multiple tough conditions.

## 5. Conclusion

Aiming at the problem that rail fastener detection mainly depends on manpower, high cost, low efficiency, and the quality of detection is easily disturbed, this paper studies the development of contemporary in-depth learning image technology, designs and implements an efficient and assured image detection method for rail fastener.

At present, most computer vision image detection technologies need labeled images for training models. There is no open data set of rail fastener image detection on the market. Previously, most of the data used by researchers are self-made, lack of uniform labeling, and may lack sufficient quantity. Therefore, this paper establishes a rail fastener image detection data set. The data set includes four types of fasteners: WJ-7, WJ-8, W300-1 and V. Each type has 1000 pictures. The dataset is annotated according to the format of PASCAL VOC detection dataset. The training set and test set are divided according to the ratio of 3:1, and the images of rail fasteners in cloudy and rainy environments are collected for robustness test.

In this paper, by replacing the basic network, data enhancement and on-line hard sample mining technology, the accuracy of the model is improved from 96.73% to 99.8%. According to the Batch-Normalization layer and Scale Layer are fused to control the number of interesting outputs of Proposal layer, and some column operations to control the size of the input image are carried out. Using TITAN X GPU, the accuracy of the model is improved from 96.73% to 99.8%. And, the speed of model has been improved to 35FPS of single GPU successfully.

The image detection method of rail fastener designed in this paper achieves good results in data set. It can be applied to actual scene, realize automatic rail fastener recognition, reduce labor cost, and improve the efficiency of railway security inspection with a broad application value.